\documentclass[11pt,a4paper]{article}
\usepackage[hyperref]{acl2019}
\usepackage{times}
\usepackage{latexsym}
\usepackage{graphicx,amsmath,amsfonts}
\usepackage{booktabs}
\usepackage{soul}

\usepackage{url}

\aclfinalcopy 
\def\aclpaperid{1156} 

\newlength{\mylength}
\title{Updating Pre-trained Word Vectors and Text Classifiers \\ using Monolingual Alignment}

\author{
Piotr Bojanowski
\quad
Onur Celebi
\quad
Tomas Mikolov
\quad
Edouard Grave
\quad
Armand Joulin
\\
Facebook AI Research
\\
\texttt{\{bojanowski,celebio,tmikolov,egrave,ajoulin\}@fb.com}
}

\date{}

\begin{document}
\maketitle

\begin{abstract}
In this paper, we focus on the problem of adapting word vector-based models to new textual data. 
Given a model pre-trained on large reference data, how can we adapt it to a smaller piece of data with a slightly different language distribution? 
We frame the adaptation problem as a monolingual word vector alignment problem, and simply average models after alignment. 
We align vectors using the RCSLS criterion. 
Our formulation results in a simple and efficient algorithm that allows adapting general-purpose models to changing word distributions. 
In our evaluation, we consider applications to word embedding and text classification models. 
We show that the proposed approach yields good performance in all setups and outperforms a baseline consisting in fine-tuning the model on new data.
\end{abstract}

\section{Introduction}
The performance of models for core NLP problems heavily rely on the availability of large amounts of high-quality training data.
Word embedding, language modeling or text classification are tasks that all benefit from training at scale.
Pre-trained models optimized on massive corpora are readily available for most tasks, and are often used in downstream applications.
For example, popular pre-trained word vectors like \texttt{word2vec} or \texttt{fastText} are trained for weeks on a large mix of web data and serve as building blocks many NLP applications.
In a similarly way text classifiers are trained offline on a fixed and large labeled training set before being shipped to an application.
A good example is the \texttt{fastText} language identifier\footnote{https://fasttext.cc/docs/en/language-identification.html} that was trained on data from Tatoeba\footnote{https://tatoeba.org} and can serve as a component of a larger system.

However, such pre-trained models are not without their flaws.
First, these general models intended for a broad range of applications suffer from the lack of specialization.
Indeed, despite their size, large web data such as Common Crawl lack coverage for highly technical expert fields such as medicine or law.
Second, many applications rely on the temporal aspect of training and test data as the language distribution can drastically change over time.
New words may appear in the vocabulary, new named entities gain sudden importance and new trends are rapidly emerging.
Because of that, in many situations general-purpose pre-trained models require adaptation to fit the distribution of the task at hand.

The simplest solution to adapt a model to specialized data is to retrain the model from scratch on the relevant data.
However, that is not always possible as it would require:
\textbf{(i)} having access to the large dataset that was used for pre-training, 
\textbf{(ii)} retaining the data history and processing an ever-growing dataset.
Another approach to adapt the model is to \emph{fine-tune} it on new data to fit the new distribution. 
This solution is technically challenging, as one has to carefully select relevant hyper parameters.
Moreoever, even when carried out carefully, it leads to a loss of important statistics gathered on the original large dataset.

In this work, we propose a simple method allowing to combine a pre-trained model with a model trained on the new data. 
We frame this problem as a word vector alignment problem and take inspiration from the recent progress made in bilingual lexicon induction. 
Our approach requires little retraining and only needs storing the previous model, not the data. 
When working with large datasets, this represents a considerable computational advantage.
We experimentally show that our approach allows to successfully adapt word vector models as well as text classifiers to new data.

\section{Problem formulation}
\label{sec:problem}
In this work, we deal with models based on word vectors, in particular, word embedding and text classification models. We suppose that we are given a model with word vectors $\mathbf{X}$ that is pre-trained on a large corpus $S_0$.
We also suppose that we have access to a novel corpus $S_1$ of limited size.
$S_1$ can differ from $S_0$ in a variety of ways, from a subtle shift in the word distribution to the appearance of new words (for instance, neologisms).
We are interested in updating the model's word vectors $\mathbf{X}$ to the specificities of the small corpus while retaining most of the information from the original vectors.
In what follows, let us denote by $V_0$ (respectively $V_1$) the lexicon found in $S_0$ (respectively $S_1$).

The classical solution to this problem is to train a new model on $S_1$ while initializing the parameters with vectors from $\mathbf{X}$.
We refer to this solution as \emph{fine-tuning}.
Two main issues arise with this approach: first, it only updates words that are in $V_1$, leaving most of the vectors in $\mathbf{X}$ untouched.
Therefore this procedure can create a large discrepancy between words in $V_1$ and $V_0 \setminus V_1$.
Second, aggressive fine-tuning may lead to loss of information from the original dataset $S_0$.
Indeed, words in $V_1$ (including those common to the two sets) will specialize in reflecting the distribution of $S_1$, discarding the useful statistics learned on $S_0$.

In this work, to adapt our model $\bf X$ to $S_1$, we propose to train a model on $S_1$, then align and average it with $\mathbf{X}$.
The advantage of this alignment-based approach is that all the word vectors are being updated.
We denote the word vectors trained on $S_1$ as $\mathbf{Y}$.
Using words in $V_0 \cap V_1$, we find a linear mapping $\mathbf{Q}$ that aligns the word vectors in $\mathbf{X}$ and $\mathbf{Y}$.
Given the mapped vectors $\mathbf{XQ}$ and $\mathbf{Y}$, we construct the final word vectors $\mathbf{Z}$ by simply taking an average:
$$
  z_i =
  \begin{cases}
    \mathbf{Q}^\top x_i & \text{if } i \in V_0 \setminus V_1, \\
    \frac{1}{2}(\mathbf{Q}^\top x_i + y_i) & \text{if } i \in V_0 \cap V_1, \\
    y_i & \text{if } i \in V_1 \setminus V_0.
  \end{cases}
$$
Please note that the same formulation allows to model the confidence in new data $S_1$ by replacing the average by a weighted sum:
$$
  (1 - \alpha) \ \mathbf{Q}^\top x_i + \alpha \ y_i,
$$
where $\alpha$ is a parameter in $[0,1]$ governing the confidence in the new data.

\section{Word vector alignment}
As described in the previous section, we are given two sets of $n$ word vectors in dimension $d$ stacked in two $n \times d$ matrices $\mathbf{X}$ and $\mathbf{Y}$.
In the case of bilingual word vector alignment, the lines of $\bf X$ and $\bf Y$ have to be put in correspondance using a bilingual lexicon.
When working with monolingual data, we assume without loss of generality that word vectors have the same index $i$ in $\mathbf{X}$ and $\mathbf{Y}$.
\citet{Mikolov13} propose to frame the word vector alignment problem as linear least squares which results in a quadratic optimization problem.
The linear mapping matrix $\mathbf{Q} \in \mathbb{R}^{d \times d}$ can be found by solving:
$$
  \min_{\bf Q} \ \frac{1}{n} \| \mathbf{XQ} - \mathbf{Y} \|_F^2,
$$
which admits a closed-form solution.
Restraining $\mathbf{Q}$ to the set of orthogonal matrices $\mathcal{O}_d$, has been shown to improve the alignments~\citep{Xing15}.
In that case, the resulting problem is known as Orthogonal Procrustes, and still admits a closed-form solution obtained using a singular value decomposition~\citep{schönemann66}.

Using an orthogonal mapping is also critical when working with classification models.
Having $\bf Q$ in $\mathcal{O}_d$ ensures that we can preserve the scoring function.
In a linear classification model the probability for sample $i$ to be of class $k$ can be written:
$$
  p(C=k | x_i) = \frac{e^{v_k^\top x_i}}{\sum_{k' = 1}^K e^{v_{k'}^\top x_i}}.
$$
We see that this probability will be unchanged if we map both features and classifiers and that the mapping $\bf Q$ is orthogonal:
$$
  v_k^\top \mathbf{Q}^\top \mathbf{Q} x_i = v_k^\top x_i.
$$

\paragraph{Alternative loss function.}
The $\ell_2$ norm described in the previous paragraphs is intrinsically associated with the euclidean nearest neighbor (NN) retrieval criterion.
This criterion suffers from the existence of ``hubs'', which are data points that are nearest neighbors to many other data points~\citep{dinu2014improving}.
To alleviate that problem, alternative criterions have been suggested in the litterature, such as the inverted softmax~\citep{smith2017offline} and CSLS~\citep{conneau2017word}.
\citet{joulin2018loss} show that minimizing a loss inspired by CSLS can significantly improve the quality of the retrieved word alignments.
Their loss function, called \textsc{RCSLS} is defined as:
\begin{align*}
  - 2 \mathbf{x}^\top \mathbf{y} + \frac{1}{k} \sum_{\mathbf{y} \in \mathcal{N}_Y (\mathbf{x})} \mathbf{x}^\top \mathbf{y} + \frac{1}{k} \sum_{\mathbf{x} \in \mathcal{N}_X (\mathbf{y})} \mathbf{x}^\top \mathbf{y}.
\label{eq:rcsls}
\end{align*}
This loss is a tight convex relaxation of the CSLS criterion for normalized word vectors.
The problem of learning an orthogonal alignment using the \textsc{RCSLS} loss can be solved using a projected sub-gradient descent method.

\section{Experiments}
We empirically show that the alignment procedure that we propose allows to succesfully update pre-trained models on new training data.
We evaluate that in three experiments: one concerning word embeddings, and two related to text classification.
In all experiments, for our approach we align $\bf X$ and $\bf Y$ using the RCSLS loss and use for $\bf Y$ vectors obtained by fine-tuning $\bf X$ on $S_1$.


\subsection{Updating word vectors}
In this first experiment, we want to check how well does our method allow to update word vector models, especially when the new corpus $S_1$ contains a lot of \emph{new words}, never seen in $S_0$.
As this kind of data is hard to find, we simulate this setup by discarding from $S_0$ lines containing selected words, that are present in $S_1$.
In order to be able to measure measure how well the update procedure works, we will create two test sets, with or without \emph{new words}.

\paragraph{Test data.}
We evaluate our word vectors on word analogies~\citep{mikolov2013efficient}.
This dataset is composed of 19544 questions grouped in 14 categories, with a vocabulary of 904 words.
In each category we select 10\% of words that we will remove from $S_0$.
This results in a set $\mathcal{B}$ composed of $104$ words.
We split the analogy dataset in two:
a first set (\emph{Out of vocab}) that only contains questions that have at least one word in $\mathcal{B}$,
and a second set (\emph{In vocab}) in which we put the leftover questions.

\paragraph{Training data.}
We take two subsets of the May 2017 dump of the Common Crawl that we preprocess following~\citet{grave2018learning}.
In order to avoid case-related problems, we lowercase the training data.
We take two subsets of imbalanced size: ($S_0$) contains 8.8 billion words, while ($S_1$) has 440 million words.
From $S_0$ we discard lines that contain at least one word in $\mathcal{B}$, yielding a dataset of 974 million words.
We train our word vectors using the \texttt{fastText} library, with no character $n$-grams, sampling 10 negatives and training the model for 10 epochs.

\begin{table}[t]
  \centering
  \begin{tabular}{l c c }
  \toprule
                  & Out of vocab & In vocab \\
  \midrule
  Train on $S_0$  & \phantom{0}0.0    & 70.1 \\
  Train on $S_1$  & 66.9 & 66.1 \\
  Train on $S_0 \cup S_1$ & 68.6 & 71.2 \\
  \midrule
  Fine-tune       & 67.7 & 66.8 \\
  Subwords        & 37.0 & 71.6 \\
  \midrule
  RCSLS+Fine.     & \bf 67.9 & \bf 72.1 \\
  \bottomrule
  \end{tabular}
  \caption{
		Performance of updated word vectors on the word analogy task.
		We split the English word analogy datasets~\citep{mikolov2013efficient} into \emph{Out of vocab} and \emph{In vocab} questions.
		\emph{Out of vocab} questions are composed of words that are out of vocabulary for $S_0$, hence the null accuracy.
	}
  \label{tab:synth}
\end{table}

\paragraph{Baselines.}
We consider two baselines:
\begin{itemize}
  \item \textbf{Fine-tune}:~train word vectors on $S_1$ and initialize the input matrix using word vectors trained on $S_0$.
  \item \textbf{Subwords}:~train on $S_0$ with subwords, with character $n$-grams of length $4$ to $6$.
  We build word vectors for all words by summing all the character $n$-gram vectors.
\end{itemize}
As a topline, we also provide the performance of a model trained on $S_0 \cup S_1$.
This variant should be considered as the oracle solution, since it requires access to the old data.

\paragraph{Results.}
The quantitative results are presented in Table~\ref{tab:synth}.
First, we see that fine-tuning $\bf X$ on $S_1$ leads to decent performance on the \emph{Out of vocab} questions, but saps the accuracy on \emph{In vocab} questions.
As mentionned before, learning vectors initialized with $\bf X$ on $S_1$ may lead to a loss of important statistics learnt on $S_0$: the total accuracy on \emph{In vocab} questions drops by $3.3\%$.

Second, we observe that training subword-enriched vectors on $S_0$ (\emph{Subwords} baseline), when compared with simple training on $S_0$, improves performance on \emph{In vocab} questions ($1.5\%$ improvement). 
However, we notice that this baseline fails to provide good features for the \emph{Out of vocab} questions, leading to a poor accuracy of $37.0 \%$.

Third, RCSLS+Fine-tune leads to best average performance for both tasks, outperforming Fine-tune and Subwords.
On \emph{In vocab} questions, our approach leads to even better accuracy than when training on $S_0 \cup S_1$ ($0.9\%$ improvement).
All in all the proposed approach allows to adapt the pre-trained model to words that were only present in $S_1$ without losing precious information from $S_0$, making it an effective method for updating word vector models.

\subsection{Updating classification models in time}
In this second experiment, we want to check if our approach allows to adapt text classification models to new data.
As opposed to the previous experiment, we perform this experiment on real data: user reviews taken at different moments in time. 
In that setup, we observe a significant change in the language distribution between data splits due to language changes over time.
A lot of named entities has changed, and the set of most discriminative words and $n$-grams for predicting sentiment may have changed too.

We focus on a linear classifier on top of unigram and bigram embeddings, and  use the \texttt{fastText} library\footnote{https://github.com/facebookresearch/fastText}\citep{joulin2016bag}.
We train models with ten hidden dimensions, and tune the number of epochs for each subset of each dataset.
Given two models trained on $S_0$ and $S_1$, we learn an orthogonal mapping between the word vectors using RCSLS.
We take as training pairs only the 1000 words with highest word-vector norms.
Using the learnt alignment, we combine word and n-gram vectors, as well as classifiers and evaluate the model on the test set.
In this experiment, we only report the Fine-tune baseline, where we train a classifier on $S_1$ and initialize the parameters with those obtained on $S_0$.

\paragraph{Dataset.}
We consider the Yelp dataset provided in their Yelp 2019 Challenge\footnote{https://www.yelp.com/dataset/challenge}.
It is a dataset composed of business reviews written by Yelp users from 2013 to 2018.
For our experiment we split the data into a large training set $S_0$ of $1.2$M reviews taken from 2013-2014, and a smaller training set $S_1$ of reviews taken from 2018.
As we want to check in this experiment what is the effect of the size of $S_1$ on the peformance of the baseline and our method, we consider four variants by growing the size of $S_1$, taking subsets of $10$k, $30$k, $100$k and $500$k samples.
We evaluate our models on a test sets composed of reviews from 2018.

\begin{table}[t!]
  \centering
  \begin{tabular}{l cccc}
    \toprule
               & \multicolumn{4}{c}{Size of $S_1$} \\
    \cmidrule{2-5} 
               & $10$k & $30$k & $100$k & $500$k \\
    \midrule
    Train on $S_0$          & 74.9 & 74.9 & 74.9 & 74.9 \\
    Train on $S_1$          & 70.8 & 72.7 & 74.4 & 76.2 \\
    Train on $S_0 \cup S_1$ & 75.1 & 75.1 & 75.3 & 76.2 \\
    \midrule
    Fine-tune               & 72.9 & 73.8 & 75.0 & \bf 76.3 \\
    \midrule
    RCSLS+Fine.   & \bf 75.1 & \bf 75.3 & \bf 75.8 & \bf 76.3 \\
    \bottomrule
  \end{tabular}
  \caption{
    Classification accuracy on Yelp reviews from 2018.
    We compare models trained on $1.2$M reviews from 2013-2014 ($S_0$) and a those trained on a smaller sample of reviews from 2018 ($S_1$).
    We vary the size of $S_1$ from $10$k to $500$k samples, while keeping $S_0$ of fixed size.
    }
  \label{tab:yelp}
\end{table}

\paragraph{Results.}
We present the result of this experiment in Table~\ref{tab:yelp}.
First of all, we observe that the performance of models trained on $S_0$ and $S_1$ strongly depend on the size of $S_1$.
When the two datasets are of the same size (500k), the best performing model is the one trained on $S_1$, as in that case there is no train/test distribution discrepancy.
However, when $S_1$ is small (10k or 30k), it is better to use a larger yet ill-distributed dataset ($70.8\%$ versus $74.9\%$).
We also observe that when training on a concatenation of the two, the model performs at least as well as the best one.

Second, for all sizes of $S_1$, our method outperforms the fine-tuning baseline.
When $S_1$ is large, it benefits from the fine-tunning, while when $S_1$ is small it takes advantage of the initial model trained on $S_0$.
A surprising observation is that in that experiment, aligning the fine-tuned vectors works as well as training on $S_0 \cup S_1$.
This shows that in some applications one can simply retain the model while discarding the old data.

\begin{table*}[t!]
  \centering
  \begin{tabular}{l c ccc c ccc c ccc}
    \toprule
               &~& \multicolumn{3}{c}{Sogou}  &~& \multicolumn{3}{c}{Amazon} &~& \multicolumn{3}{c}{Yelp}\\
    \cmidrule{3-5} \cmidrule{7-9}\cmidrule{11-13}
               && $3$k & $30$k & Full     &  & $3$k & $30$k & Full        &  & $3$k & $30$k & Full \\
    \midrule
    Train on $S_0$            && 91.3 & 94.5 & 96.2 && 46.1 & 53.2 & 59.5 && 50.7 & 58.7 & 62.8 \\ 
    Train on $S_1$            && 91.4 & 94.3 & 96.1 && 47.2 & 53.8 & 59.6 && 50.2 & 59.0 & 62.8 \\
    Train on $S_0 \cup S_1$   && 92.3 & 95.1 & 96.7 && 48.6 & 54.9 & 60.2 && 54.2 & 60.1 & 63.7 \\
    \midrule                          
    Fine-tune                 && 91.5 & 94.3 & 96.1 && 47.2 & 53.8 & 59.6 && 52.4 & 59.0 & 62.9 \\
    Vote                      && \bf 92.0 & \bf 94.8 & \bf 96.3 && 48.1 & 54.3 & 59.9 && 51.1 & 59.8 & 63.9 \\
    \midrule
    RCSLS+Fine.               && \bf 92.0 & \bf 94.8 & 96.2 && \bf 48.2 & \bf 54.4 & \bf 60.0 && \bf 52.7 & \bf 60.2 & \bf 64.0 \\
    \bottomrule
  \end{tabular}
  \caption{
    Comparison between the accuracy of our aligned model (\emph{RCSLS+Fine.}), the accuracy of models trained on each split separately (\emph{Train on \dots}) and two baselines. 
    We observe that a model obtained using our procedure matches the performance of the \emph{Vote} baseline while not requiring to store two separate models.
  }
  \label{tab:sup}
\end{table*}


\subsection{Merging text classification models}
\label{sec:zhang}
In this final experiment, we want to evaluate how our approach compares to a standard technique of model ensembeling.
To this end, we perform a control experiment in which we evaluate how well we can combine models trained on two shards of data of similar size.
A standard technique for making an ensemble of classification models is voting, or averaging the output of the scoring function. 

In the case of linear models, averaging the output of the scoring function is exactly equivalent to averaging the parameters.
However, classification models such as the one used by~\citet{joulin2016bag} are based on a low rank parametrization of the classifier.
Because of that, directly averaging the parameters is not possible:
$$
  {\bf W}_0 + {\bf W}_1 = {\bf X}_0 {\bf V}_0^\top + {\bf X}_1 {\bf V}_1^\top.
$$
By finding an orthogonal matrix $\bf Q$ that maps ${\bf X}_0$ to ${\bf X}_1$, we should be able to do so:
$$
  {\bf W}_0 + {\bf W}_1 \approx \frac{1}{2} \left ( {\bf X}_0 {\bf Q} + {\bf X}_1 \right ) \left ( {\bf V}_0 {\bf Q} + {\bf V}_1 \right )^\top.  
$$
For all classifiers trained in this experiment, we follow the same procedure as in the previous experiment.
We train a \texttt{fastText} model with bi-grams and tune the number of epochs on the validation set.

\paragraph{Datasets.}
For this experiment, we use a subset of the datasets proposed by~\citet{zhang2015character}, i.e., Sogou News, Amazon and Yelp full reviews.
We randomly split each dataset into two subsets $S_0$ and $S_1$ of increasing size: $3$k samples, $30$k samples and up to the full dataset.
For any split and any method, we test our classifier on the full corresponding test set.

\paragraph{Baselines.}
As in the previous experiments, we report a \emph{Fine-tune} baseline.
To this end, when training a model on $S_1$, we initialize the input matrix with word and $n$-gram embeddings obtained when learning a classifier on $S_0$.
In that case, the classifiers are initialized randomly.

We also report the performance obtained by training a model on each of the two splits and then aggregating the predictions by voting.
Since we only use two models, in case of disagreement we take the most confident prediction.
For the reasons exposed at the beginning of this section, this baseline should perform the same as our method.

\paragraph{Quantitative results.}
We report the performance of a model trained on each of the two splits alone, the two baselines, and a topline obtained by training on $S_0 \cup S_1$.
The quantitative results are presented in Table~\ref{tab:sup}.
First of all, we observe that our approach reduces the gap between training on a single set with the topline.
This effect is especially true on small versions of the datasets ($3$k or $30$k).
The main reason for this improvement is that each split has an incomplete coverage of the discriminative words and $n$-grams.

Second, and most importantly, we notice that our approach performs comparably to the \emph{Vote} baseline, which validates our claims.
By aligning the word vectors and averaging the models, we manage to get the performance of a model ensemble, while only storing a single model.

\section{Conclusion}
We presented a simple method to update word vectors to the distribution of a new corpus.
Our method is not a definitive solution to this challenging task, rather constitutes a proof of concept.
Experiments seem to indicate that the proposed approach could be used for extending the lexicon, allowing to aggregate low frequency words from several corpora.
While this task is of premier importance, we lack proper evaluation datasets for rare words.
We leave the construction of adapted evaluation datasets for future work, and posit that such resources would greatly fuel research in that direction.

\bibliography{acl2019}
\bibliographystyle{acl_natbib}

\end{document}